\documentclass[10pt,twocolumn,letterpaper]{article}
\usepackage{cvpr,times}
\usepackage{url}
\pdfoutput=1

\usepackage{bm}
\usepackage{latexsym}
\usepackage{amsmath}
\usepackage{amssymb}
\usepackage{booktabs}
\usepackage{epsfig}
\usepackage{graphicx}
\usepackage{xspace}
\usepackage{wrapfig}
\usepackage{caption}
\usepackage{subcaption}
\usepackage{mathtools}
\usepackage[pagebackref=true,breaklinks=true,letterpaper=true,colorlinks,bookmarks=false]{hyperref}

\cvprfinalcopy 

\def\cvprPaperID{1448} 

\DeclarePairedDelimiter\norm{\lVert}{\rVert}%
\makeatletter
\let\oldnorm\norm
\def\norm{\@ifstar{\oldnorm}{\oldnorm*}}
\makeatother

\ifcvprfinal\fi

\title{Deep Domain Confusion: Maximizing for Domain Invariance}

\author{
Eric Tzeng, Judy Hoffman, Ning Zhang \\
UC Berkeley, EECS \& ICSI\\
\footnotesize{\texttt{\{etzeng,jhoffman,nzhang\}@eecs.berkeley.edu} }\\
\and
Kate Saenko \\
UMass Lowell, CS  \\
\footnotesize{\texttt{saenko@cs.uml.edu}} \\
\and
Trevor Darrell \\
UC Berkeley, EECS \& ICSI\\
\footnotesize{\texttt{trevor@eecs.berkeley.edu}} \\
}

\newcommand{\daume}{Daum\'e~III\xspace}

\begin{document}

\maketitle

\begin{abstract}
Recent reports suggest that a generic supervised deep CNN model trained on a
large-scale dataset reduces, but does not remove, dataset bias on a standard
benchmark. Fine-tuning deep models in a new domain can require a significant
amount of data, which for many applications is simply not available.  We propose
a new CNN architecture which introduces an adaptation layer and an additional
domain confusion loss, to learn a representation that is both semantically
meaningful and domain invariant. We additionally show that a domain confusion
metric can be used for model selection to determine the dimension of an
adaptation layer and the best position for the layer in the CNN architecture.
Our proposed adaptation method offers empirical performance which exceeds
previously published results on a standard benchmark visual domain adaptation
task.

\end{abstract}

\section{Introduction}
Dataset bias is a well known problem with traditional supervised approaches to image
recognition~\cite{efros-cvpr11}.  A number of recent theoretical and empirical 
results have shown that supervised methods' test error increases in
proportion to the difference between the test and training input
distribution \cite{ben2007analysis, blitzer2007learning,saenko-eccv10,efros-cvpr11}.  In the last few
years several methods for visual domain adaptation have been suggested
to overcome this
issue \cite{daume,yang-icdm07,aytar-iccv11,saenko-eccv10,kulis-cvpr11,Khosla-eccv12,gopalan-iccv11,gong-cvpr12,hoffman-eccv12,hoffman-iclr13},
but were limited to shallow models. The traditional approach to
adapting deep models has been fine-tuning; see \cite{rcnn} for a
recent example.

Directly fine-tuning a deep network's parameters on a small amount of labeled
target data turns out to be problematic.  
Fortunately, pre-trained deep models do perform well in novel domains.
Recently, \cite{decaf,hoffman-iclr14} showed that using the deep
mid-level features learned on ImageNet, instead of the more
conventional bag-of-words features, effectively removed the bias in
some of the domain adaptation settings in the Office
dataset \cite{saenko-eccv10}.   These algorithms transferred
the representation from a large scale domain, ImageNet, as well as using all of the data in that domain as source data for 
appropriate categories.  
However, these methods have no way to select a representation from the deep architecture and instead report results across multiple layer selection choices. 

\begin{figure}
\centering
\includegraphics[width=.8\linewidth]{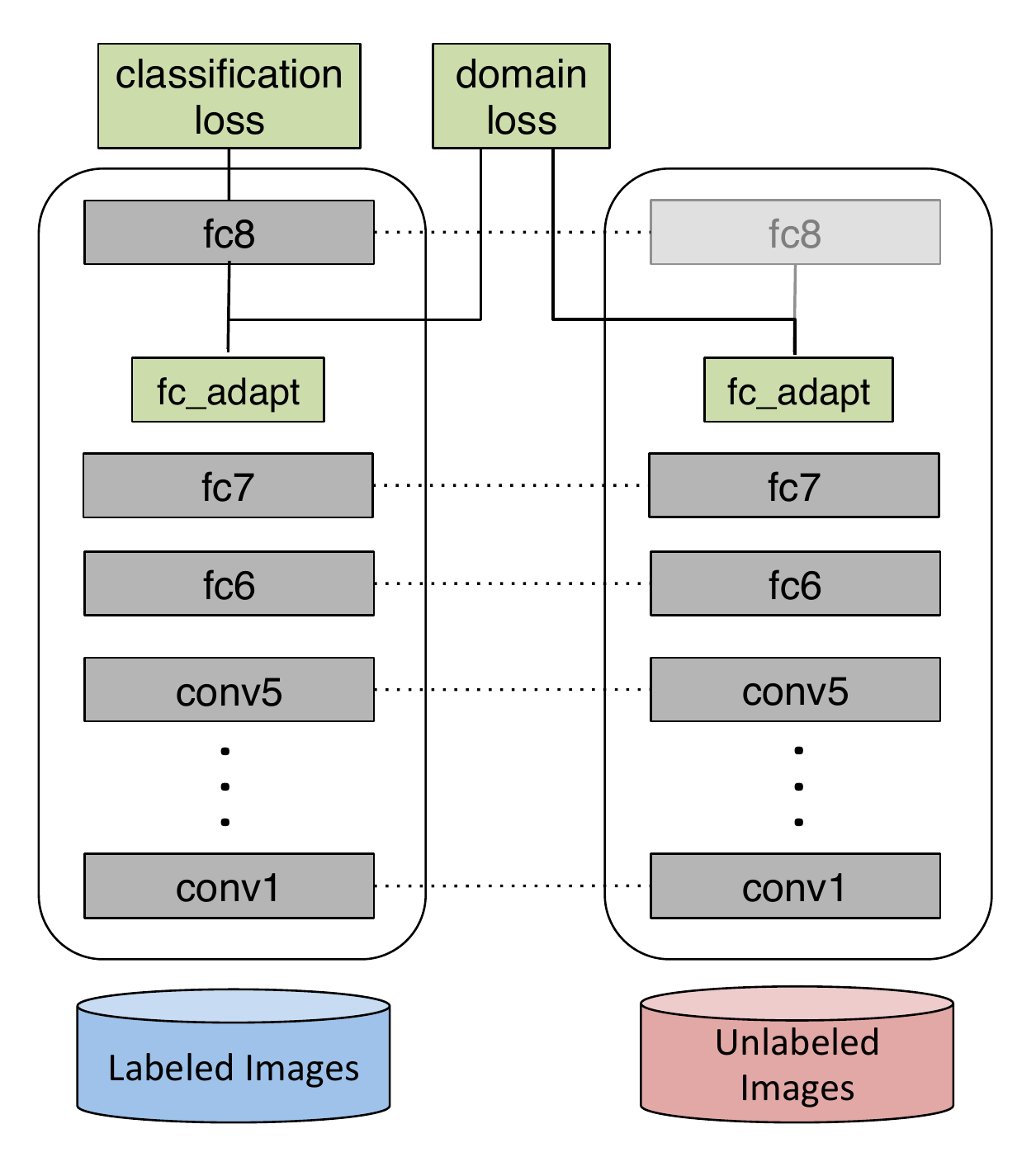}
\caption{Our architecture optimizes a deep CNN for both classification loss as
  well as domain invariance. The model can be trained for \emph{supervised}
  adaptation, when there is a small amount of target labels available, or
  \emph{unsupervised} adaptation, when no target labels are available.  We
  introduce domain invariance through \emph{domain confusion} guided selection
  of the depth and width of the adaptation layer, as well as an additional
  domain loss term during fine-tuning that directly minimizes the distance
  between source and target representations.}
\label{fig:architecture}
\end{figure}

Dataset bias was classically illustrated in computer vision by way of the ``name
the dataset'' game of Torralba and Efros~\cite{efros-cvpr11}. Indeed, this turns
out to be formally connected to measures of domain discrepancy~\cite{adist,mmd}.
Optimizing for domain invariance, therefore, can be considered equivalent to the
task of learning to predict the class labels while simultaneously finding a
representation that makes the domains appear as similar as possible. This
principle forms the essence of our proposed approach. We learn deep representations
by optimizing over a loss which includes both classification error on the
labeled data as well as a \emph{domain confusion} loss which seeks to make the
domains indistinguishable.

We propose a new CNN architecture, outlined in Figure~\ref{fig:architecture},
which uses an adaptation layer along with a domain confusion loss based on
maximum mean discrepancy (MMD)~\cite{mmd} to automatically learn a
representation jointly trained to optimize for classification and domain
invariance. We show that our domain confusion metric can be used both to select
the dimension of the adaptation layers, choose an effective placement for a new
adaptation layer within a pre-trained CNN architecture, and fine-tune the
representation.

Our architecture can be used to solve both \emph{supervised adaptation}, when a
small amount of target labeled data is available, and \emph{unsupervised
adaptation}, when no labeled target training data is available. We provide a
comprehensive evaluation on the popular Office benchmark for classification
across visually distinct domains~\cite{saenko-eccv10}. We demonstrate that by
jointly optimizing for domain confusion and classification, we are able to
significantly outperform the current state-of-the-art visual domain adaptation
results. In fact, for the case of minor pose, resolution, and lighting changes,
our algorithm is able to achieve 96\% accuracy on the target domain,
demonstrating that we have in fact learned a representation that is invariant to
these biases.

\section{Related work}
The concept of visual dataset bias was popularized in~\cite{efros-cvpr11}. There
have been many approaches proposed in recent years to solve the visual domain
adaptation problem. All recognize that there is a shift in the distribution of
the source and target data representations. In fact, the size of a domain shift
is often measured by the distance between the source and target subspace
representations~\cite{mmd,sa,adist, disc,tca}. A large number of methods have
sought to overcome this difference by learning a feature space transformation to
align the source and target representations~\cite{saenko-eccv10,kulis-cvpr11,
sa, gong-cvpr12}. For the \emph{supervised} adaptation scenario, when a limited
amount of labeled data is available in the target domain, some approaches have
been proposed to learn a target classifier regularized against the source
classifier~\cite{yang-icdm07, aytar-iccv11, BergamoTorresani10}. Others have
sought to both learn a feature transformation and regularize a target classifier
simultaneously~\cite{hoffman-iclr13, duan-icml12}.

Recently, supervised convolutional neural network (CNN) based feature
representations have been shown to be extremely effective for a variety of
visual recognition tasks~\cite{supervision,decaf, rcnn, overfeat}.  In
particular, using deep representations dramatically reduce the effect of
resolution and lighting on domain shifts~\cite{decaf, hoffman-iclr14}.

Parallel CNN architectures such as Siamese networks have been shown to be
effective for learning invariant representations~\cite{bromley1993signature,
chopra2005learning}. However, training these networks requires labels for each
training instance, so it is unclear how to extend these methods to unsupervised
settings.

Multimodal deep learning architectures have also been explored to learn
representations that are invariant to different input
modalities~\cite{ngiam2011multimodal}. However, this method operated primarily
in a generative context and therefore did not leverage the full representational
power of supervised CNN representations.

Training a joint source and target CNN architecture was proposed
by~\cite{ref:dlid}, but was limited to two layers and so was significantly
outperformed by the methods which used a deeper architecture~\cite{supervision},
pre-trained on a large auxiliary data source (ex: ImageNet~\cite{ilsvrc2012}).

\cite{da-mmd} proposed pre-training with a denoising auto encoder, then training a two-layer network
simultaneously with the MMD domain confusion loss. This effectively learns a
domain invariant representation, but again, because the learned network is
relatively shallow, it lacks the strong semantic representation that is learned
by directly optimizing a classification objective with a supervised deep CNN.


\section{Training CNN-based domain invariant representations}
\label{sec:method}
\begin{figure}
\centering
\includegraphics[width=\linewidth]{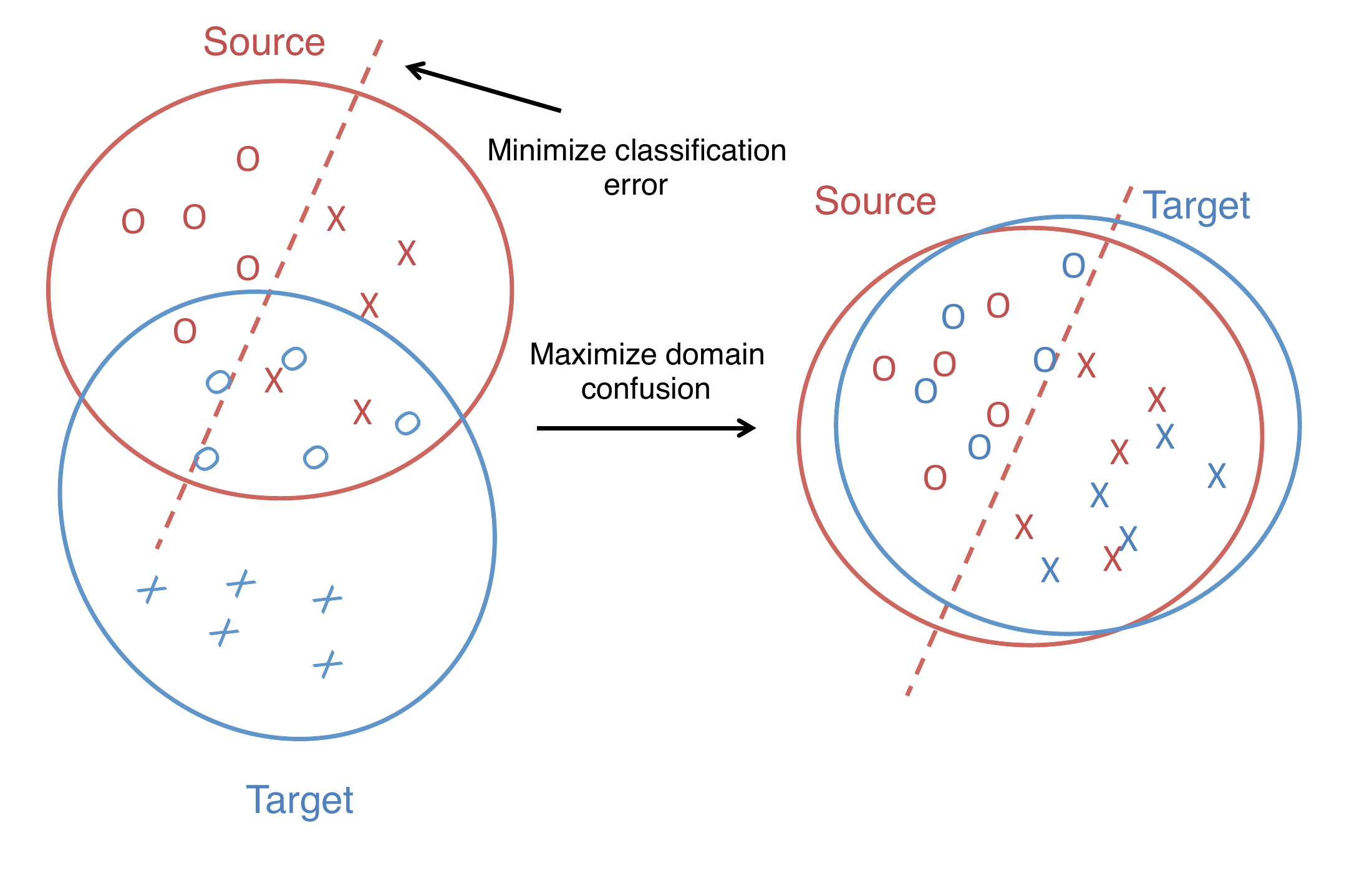}
\caption{For biased datasets (left), classifiers learned in a source domain do
not necessarily transfer well to target domains. By optimizing an objective
that simultaneously minimizes classification error and maximizes \emph{domain
confusion} (right), we can learn representations that are discriminative and
domain invariant.}

\label{fig:domain-confusion}
\end{figure}

We introduce a new convolutional neural network (CNN) architecture which we use to learn a visual representation that is both domain invariant and which offers strong semantic separation. It has been shown that a pre-trained CNN can be adapted for a new task through fine-tuning~\cite{rcnn,overfeat, lsda}. However, in the domain adaptation scenario there is little, or no, labeled training data in the target domain so we can not directly fine-tune for the categories of interest, $C$ in the target domain, $T$. Instead, we will use data from a related, but distinct  source domain, $S$, where more labeled data is available from the corresponding categories, $C$.

Directly training a classifier using only the source data often leads to overfitting to the source distribution, causing reduced performance at test time when recognizing in the target domain. Our intuition is that if we can learn a representation that minimizes the distance between the source and target distributions, then we can train a classifier on the source labeled data and directly apply it to the target domain with minimal loss in accuracy. 

To minimize this distance, we consider the standard distribution distance metric, Maximum Mean Discrepancy (MMD)~\cite{mmd}. This distance is computed with respect to a particular representation, $\phi(\cdot)$. In our case, we define a representation, $\phi(\cdot)$, which operates on source data points, $x_s \in X_S$, and target data points, $x_t \in X_T$.  Then an empirical approximation to this distance is computed as followed:
\begin{multline}
  \text{MMD}(X_S, X_T) = \\ \norm{\frac{1}{|X_S|} \sum_{x_s \in X_S} \phi(x_s) - \frac{1}{|X_T|} \sum_{x_t \in X_T} \phi(x_t)}
\end{multline}

As Figure~\ref{fig:domain-confusion} shows, not only do we want to minimize the
distance between domains (or maximize the \emph{domain confusion}), but we want
a representation which is conducive to training strong classifiers. Such a
representation would enable us to learn strong classifiers that readily transfer
across domains. One approach to meeting both these criteria is to minimize the
loss:
\begin{equation}
  \mathcal{L} = \mathcal{L}_C(X_L, y) + \lambda \text{MMD}^2(X_S, X_T)
  \label{eqn:objective}
\end{equation}
where $\mathcal{L}_C(X_L,y)$ denotes classification loss on the available
labeled data, $X_L$, and the ground truth labels, $y$, and MMD$(X_S, X_T)$
denotes the distance between the source data, $X_S$, and the target data, $X_T$.
The hyperparameter $\lambda$ determines how strongly we would like to confuse
the domains.

One approach to minimizing this loss is to take a fixed CNN, which is already a
strong classification representation, and use MMD to decide which layer to use
activations from to minimize the domain distribution distance. We can then use
this representation to train another classifier for the classes we are
interested in recognizing. This can be viewed as coordinate descent on
Eqn.~\ref{eqn:objective}: we take a network that was trained to minimize
$\mathcal{L}_C$, select the representation that minimizes MMD, then use that
representation to again minimize $\mathcal{L}_C$.

However, this approach is limited in that it cannot directly adapt the
representation---instead, it is constrained to selecting from a set of fixed
representations.  Thus, we propose creating a network to directly optimize the
classification and domain confusion objectives, outlined in
Figure~\ref{fig:architecture}.

We begin with the Krizhevsky architecture~\cite{supervision}, which has five
convolutional and pooling layers and three fully connected layers with
dimensions \{4096, 4096, $|C|$\}. We additionally add a lower dimensional,
``bottleneck,'' adaptation layer. Our intuition is that a lower dimensional layer
can be used to regularize the training of the source classifier and prevent
overfitting to the particular nuances of the source distribution. We place the
domain distance loss on top of the ``bottleneck" layer to directly regularize
the representation to be invariant to the source and target domains.

There are two model selection choices that must be made to add our adaptation
layer and the domain distance loss.  We must choose where in the network to
place the adaptation layer and we must choose the dimension of the layer.  We
use the MMD metric to make both of these decisions. First, as previously
discussed, for our initial fixed representation we find the layer which
minimizes the empirical MMD distance between all available source and target
data, in our experiments this corresponded to placing the layer after the fully
connected layer, $fc7$.

Next, we must determine the dimension for our adaptation layer. We solve this
problem with a grid search, where we fine-tune multiple networks using various
dimensions and compute the MMD in the new lower dimension representation,
finally choosing the dimension which minimizes the source and target distance.

Both the selection of which layer's representation to use (``depth'') and how
large the adaptation layer should be (``width'') are guided by MMD, and thus can
be seen as descent steps on our overall objective.

Our architecture (see Figure~\ref{fig:architecture}) consists of a source and target
CNN, with shared weights. Only the labeled examples are used to compute the
classification loss, while all data is used from both domains to compute the
domain confusion loss. The network is jointly trained on all available source
and target data.

The objective outlined in Eqn.~\ref{eqn:objective} is easily represented by this
convolutional neural network where MMD is computed over minibatches of source
and target data. We simply use a fork at the top of the network, after the
adaptation layer. One branch uses the labeled data and trains a classifier, and
the other branch uses all the data and computes MMD between source and target.

After fine-tuning this architecture, owing to the two terms in the joint loss,
the adaptation layer learns a representation that can effectively discriminate
between the classes in question due to the classification loss term, while still
remaining invariant to domain shift due the MMD term. We expect that such a
representation will thus enable increased adaptation performance.

\section{Evaluation}
\label{sec:eval}

We evaluate our adaptation algorithm on a standard domain adaptation dataset
with small-scale source domains.  We show that our algorithm is effectively able
to adapt a deep CNN representation to a target domain with limited or no target
labeled data.

The Office~\cite{saenko-eccv10} dataset is a collection of images from three
distinct domains: Amazon, DSLR, and Webcam. The 31 categories in the dataset
consist of objects commonly encountered in office settings, such as keyboards,
file cabinets, and laptops. The largest domain has 2817 labeled images.

We evaluate our method across 5 random train/test splits for each of the 3
transfer tasks commonly used for evaluation (Amazon$\rightarrow$Webcam,
DSLR$\rightarrow$Webcam, and Webcam$\rightarrow$DSLR) and report averages and
standard errors for each setting. We compare in both supervised and unsupervised
scenarios against the numbers reported by six recently published methods.

We follow the standard training protocol for this dataset of using 20 source
examples per category for the Amazon source domain and 8 images per category for
Webcam or DSLR as the source domains~\cite{saenko-eccv10,gong-cvpr12}. For the
supervised adaptation setting we assume 3 labeled target examples per category.

\subsection{Evaluating adaptation layer placement}

\begin{figure}
\centering
\includegraphics[width=\linewidth]{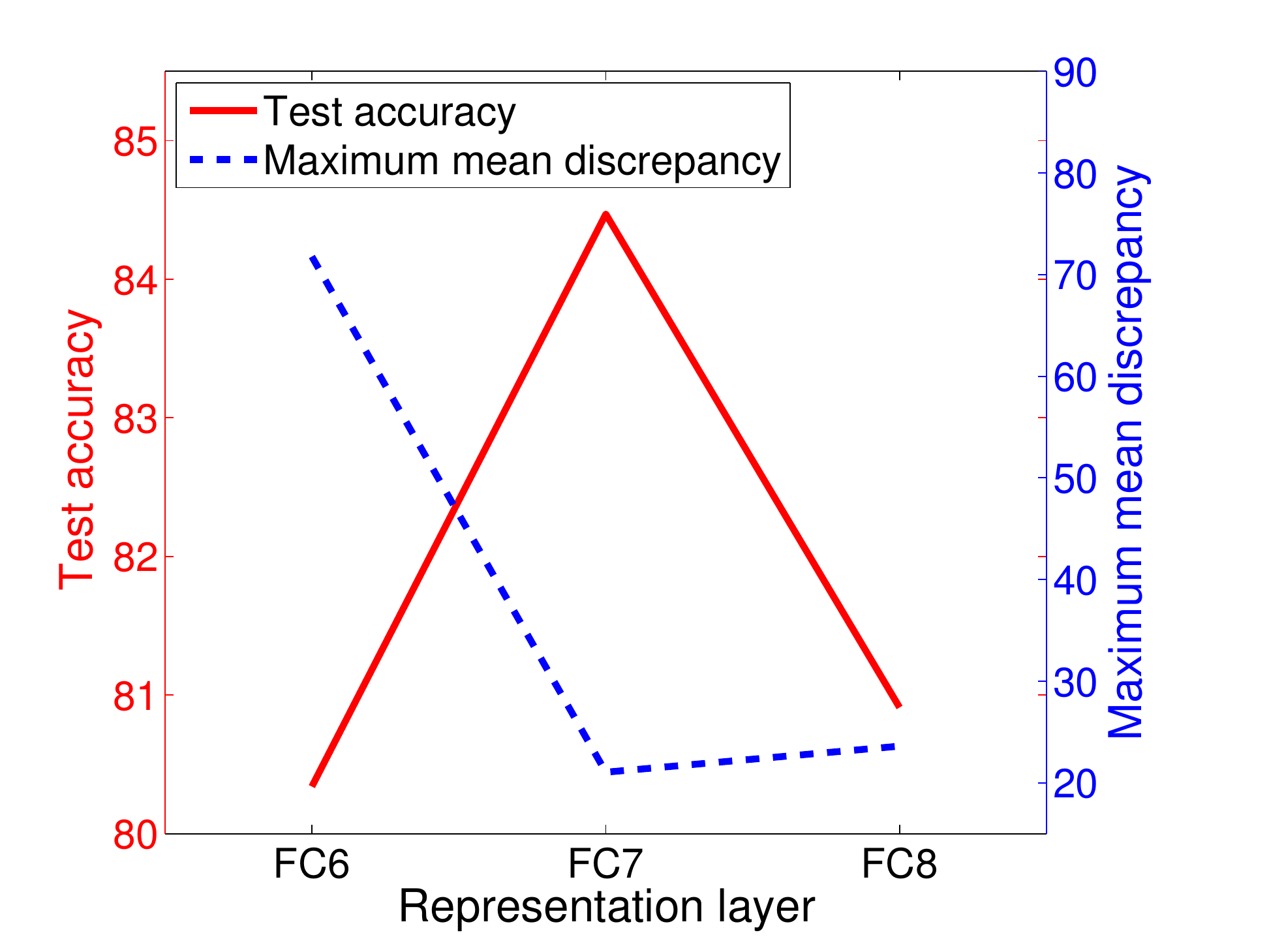}
\caption{Maximum mean discrepancy and test accuracy for different choices of
representation layer. We observe that MMD between source and target
and accuracy on the target domain test set seem inversely related, indicating
that MMD can be used to help select a layer for adaptation.}
\label{fig:mmd-depth}
\end{figure}

We begin with an evaluation of our representation selection strategy. Using a
pre-trained convolutional neural network, we extract features from source and
target data using the representations at each fully connected layer. We can then
compute the MMD between source and target at each layer. Since a lower MMD
indicates that the representation is more domain invariant, we expect the
representation with the lowest MMD to achieve the highest performance after
adaptation.

To test this hypothesis, for one of the Amazon$\rightarrow$Webcam splits we
apply a simple domain adaptation baseline introduced by \daume~\cite{daume} to
compute test accuracy for the target domain. Figure~\ref{fig:mmd-depth} shows a
comparison of MMD and adaptation performance across different choices of bridge
layers. We see that MMD correctly ranks the representations, singling out $fc7$
as the best performing layer and $fc6$ as the worst. 
Therefore, we add our adaptation layer after $fc7$ for the remaining experiments.

\subsection{Choosing the adaptation layer dimension}

\begin{figure}
\centering
\includegraphics[width=\linewidth]{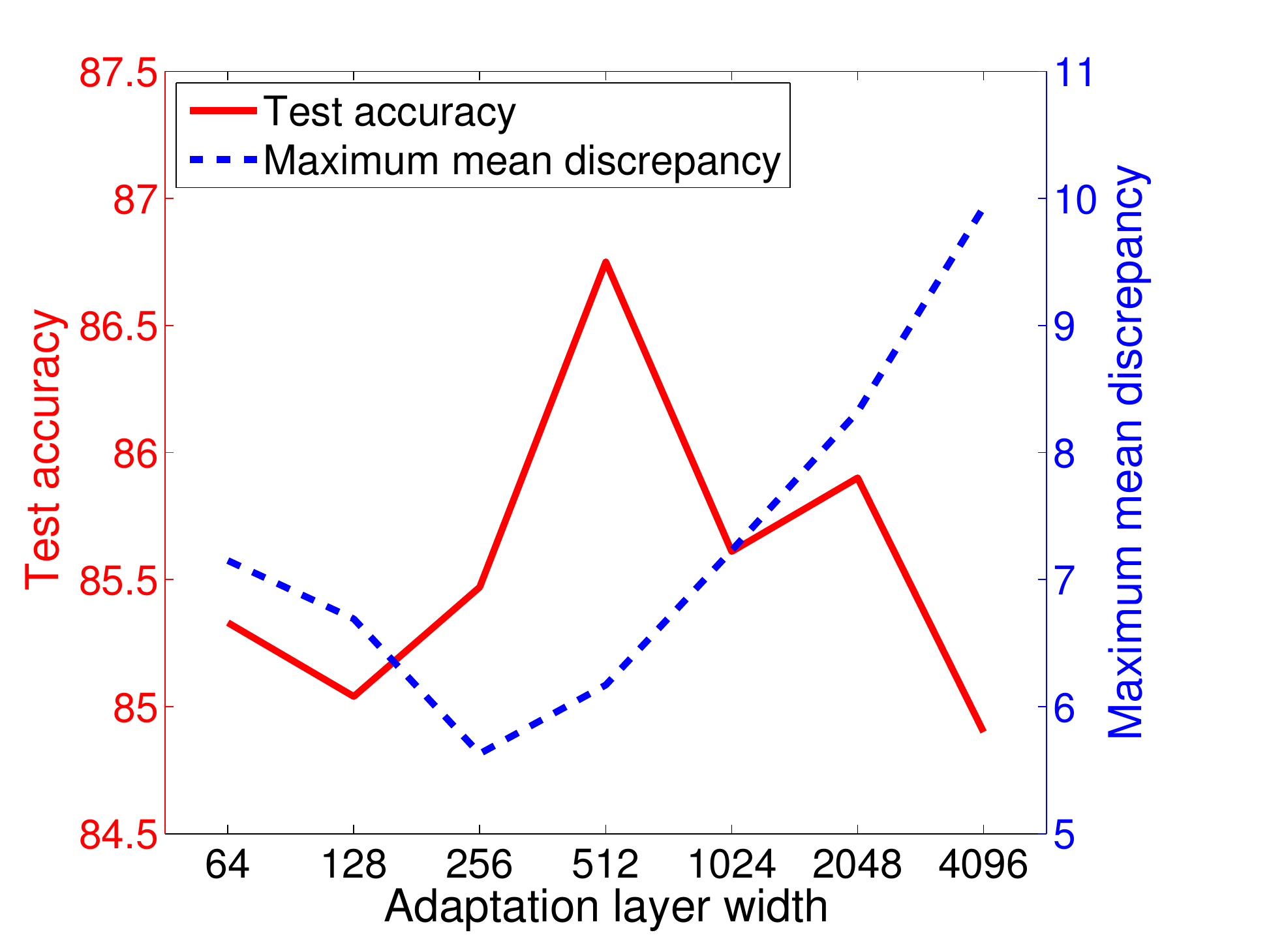}
\caption{Maximum mean discrepancy and test accuracy for different values of
adaptation layer dimensionality. We observe that MMD between source and target
and accuracy on the target domain test set seem inversely related, indicating
that MMD can be used to help select a dimensionality to use.}
\label{fig:mmd-width}
\end{figure}

\begin{table*}
  \setlength{\tabcolsep}{4pt}
\centering
\begin{tabular}{lccc|c}
\toprule
                                 & $A \rightarrow W$  & $D \rightarrow W$  & $W \rightarrow D$  & Average  \\
\midrule
GFK(PLS,PCA)~\cite{gong-cvpr12}  & 46.4 $\pm$ 0.5     & 61.3 $\pm$ 0.4     & 66.3 $\pm$ 0.4     & 53.0 \\
SA~\cite{sa}                     & 45.0               & 64.8               & 69.9               & 59.9 \\
DA-NBNN~\cite{da-nbnn}           & 52.8 $\pm$ 3.7     & 76.6 $\pm$ 1.7     & 76.2 $\pm$ 2.5     & 68.5 \\
DLID~\cite{ref:dlid}             & 51.9               & 78.2               & 89.9               & 73.3 \\
DeCAF$_6$ S+T~\cite{decaf}       & 80.7 $\pm$ 2.3     & 94.8 $\pm$ 1.2     & --                 &  --  \\
DaNN~\cite{da-mmd}               & 53.6 $\pm$ 0.2     & 71.2 $\pm$ 0.0     & 83.5 $\pm$ 0.0     & 69.4 \\
\midrule
Ours                             & \textbf{84.1 $\pm$ 0.6} & \textbf{95.4 $\pm$ 0.4} & \textbf{96.3 $\pm$ 0.3}  & \textbf{91.9} \\
\bottomrule
\end{tabular}

\caption{Multi-class accuracy evaluation on the standard supervised adaptation
  setting with the \emph{Office} dataset.  We evaluate on all 31 categories
  using the standard experimental protocol from \cite{saenko-eccv10}. Here, we
  compare against six state-of-the-art domain adaptation methods.}
\label{table:full-semi}
\end{table*}

\begin{table*}
  \setlength{\tabcolsep}{4pt}
\centering
\begin{tabular}{lccc|c}
\toprule
                                        & $A \rightarrow W$  & $D \rightarrow W$  & $W \rightarrow D$  & Average  \\
\midrule
GFK(PLS,PCA)~\cite{gong-cvpr12}         & 15.0 $\pm$ 0.4     & 44.6 $\pm$ 0.3     & 49.7 $\pm$ 0.5     & 36.4 \\
SA~\cite{sa}                            & 15.3               & 50.1               & 56.9               & 40.8 \\
DA-NBNN~\cite{da-nbnn}                  & 23.3 $\pm$ 2.7     & 67.2 $\pm$ 1.9     & 67.4 $\pm$ 3.0     & 52.6 \\
DLID~\cite{ref:dlid}                    & 26.1               & 68.9               & 84.9               & 60.0 \\
DeCAF$_6$ S~\cite{decaf}                & 52.2 $\pm$ 1.7     & 91.5 $\pm$ 1.5     & --                 &  --  \\
DaNN~\cite{da-mmd}                      & 35.0 $\pm$ 0.2     & 70.5 $\pm$ 0.0     & 74.3 $\pm$ 0.0     & 59.9 \\
\midrule
Ours                                    & \textbf{59.4 $\pm$ 0.8} & \textbf{92.5 $\pm$ 0.3} & \textbf{91.7 $\pm$ 0.8}    & \textbf{81.2} \\
\bottomrule
\end{tabular}

\caption{Multi-class accuracy evaluation on the standard unsupervised adaptation
  setting with the \emph{Office} dataset.  We evaluate on all 31 categories
  using the standard experimental protocol from \cite{gong-cvpr12}. Here, we
  compare against six state-of-the-art domain adaptation methods.}
\label{table:full-unsuper}
\end{table*}

Before we can learn a new representation via our proposed fine-tuning method, we
must determine how wide this representation should be. Again, we use MMD as the
deciding metric.

In order to determine what dimensionality our learned adaptation layer should
have, we train a variety of networks with different widths on the
Amazon$\rightarrow$Webcam task, as this is the most challenging of the three. In
particular, we try different widths varying from 64 to 4096, stepping by a power
of two each time.  Once the networks are trained, we then compute MMD between
source and target for each of the learned representations. Our method then
selects the dimensionality that minimizes the MMD between the source and target
data.

To verify that MMD makes the right choice, again we compare MMD with performance
on a test set.  Figure~\ref{fig:mmd-width} shows that we select 256 dimensions
for the adaptation layer, and although this setting is not the one that
maximizes test performance, it appears to be a reasonable choice. In particular,
using MMD avoids choosing either extreme, near which performance suffers. It is
worth noting that the plot has quite a few irregularities---perhaps finer
sampling would allow for a more accurate choice.

\subsection{Fine-tuning with domain confusion regularization}
\begin{figure*}[t]
\centering
\includegraphics[width=0.7\linewidth]{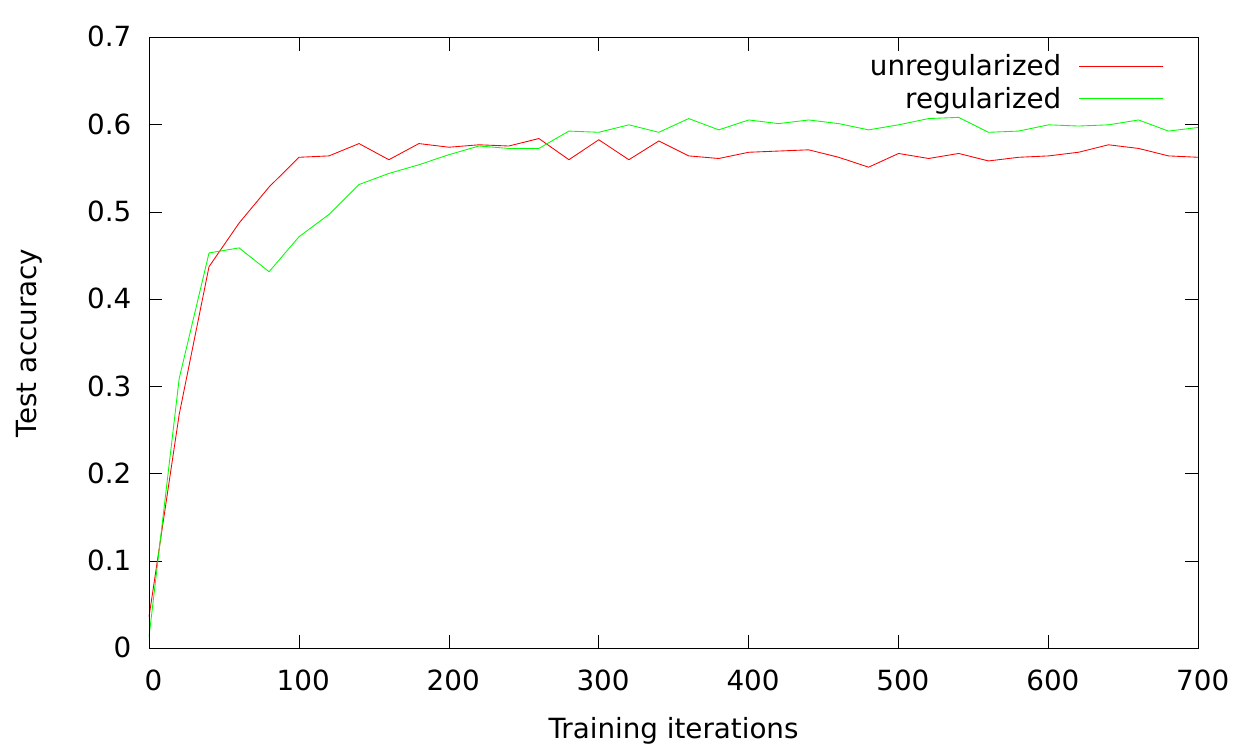}
\caption{A plot of the test accuracy on an unsupervised
Amazon$\rightarrow$Webcam split during the first 700 iterations of fine-tuning
for both regularized and unregularized methods. Although initially the
unregularized training achieves better performance, it overfits to the source
data. In contrast, using regularization prevents overfitting, so although
initial learning is slower we ultimately see better final performance.}
\label{fig:learning-curve}
\end{figure*}


Once we have settled on our choice of adaptation layer dimensionality, we can
begin fine-tuning using the joint loss described in
Section~\ref{sec:method}. However, we need to set the regularization
hyperparameter $\lambda$. Setting $\lambda$ too low will cause the MMD
regularizer have no effect on the learned representation, but setting $\lambda$
too high will regularize too heavily and learn a degenerate representation in
which all points are too close together. We set the regularization
hyperparameter to $\lambda=0.25$, which makes the objective primarily weighted
towards classification, but with enough regularization to avoid overfitting.

We use the same fine-tuning architecture for both unsupervised and
supervised. However, in the supervised setting, the classifier is trained on
data from both domains, whereas in the unsupervised setting, due to the lack of
labeled training data, the classifier sees only source data. In both settings,
the MMD regularizer sees all of the data, since it does not require labels.

Finally, because the adaptation layer and classifier are being trained from
scratch, we set their learning rates to be 10 times higher than the lower layers
of the network that were copied from the pre-trained model. Fine-tuning then
proceeds via standard backpropagation optimization.

The supervised adaptation setting results are shown in
Table~\ref{table:full-semi} and the unsupervised adaptation results are shown in
Table~\ref{table:full-unsuper}. We notice that our algorithm dramatically
outperforms all of the competing methods. The distinct improvement of our method
demonstrates that the adaptation layer learned via MMD regularized fine-tuning
is able to succesfully transfer to a new target domain.

\begin{figure*}[p]
\includegraphics[width=\textwidth]{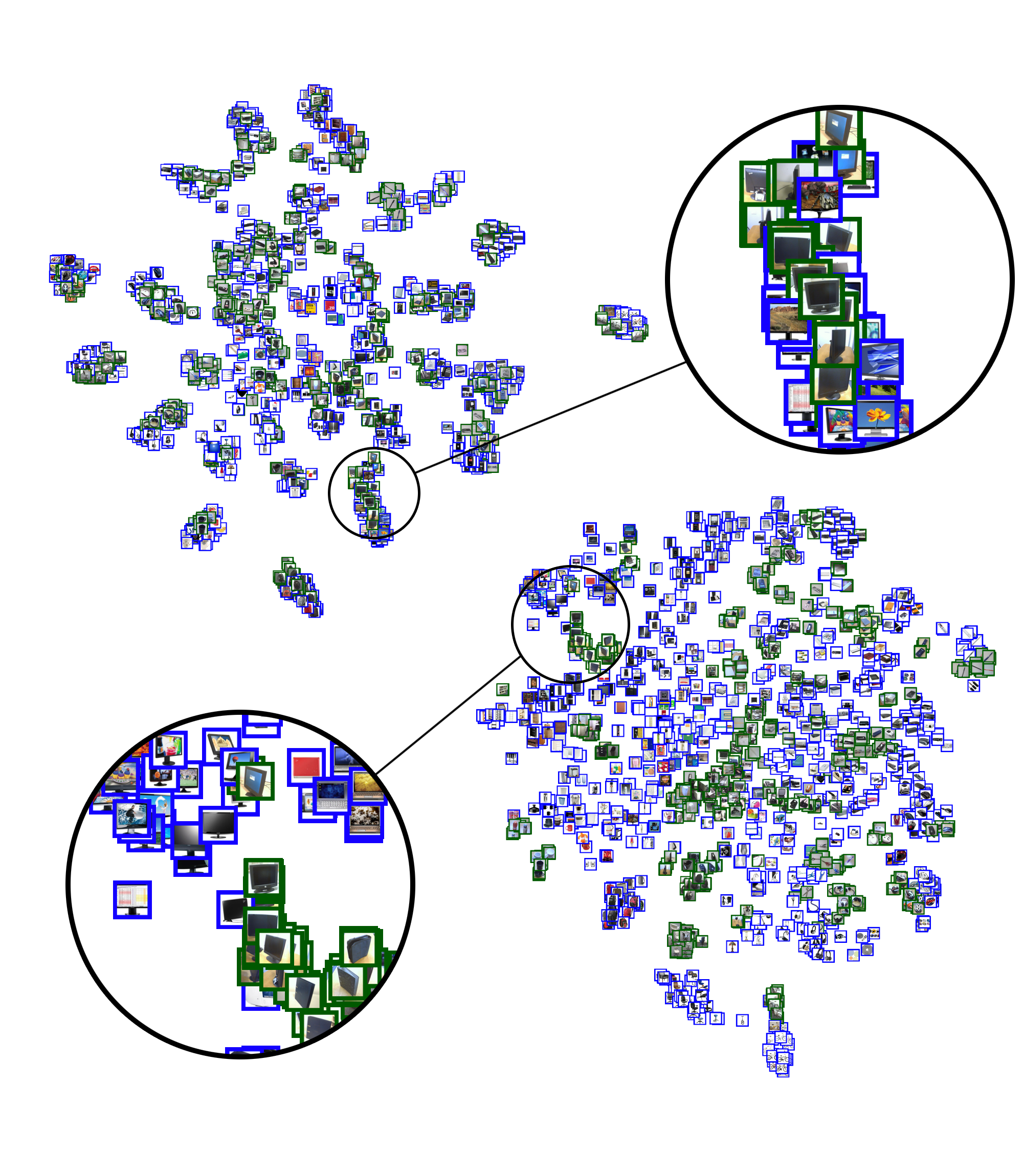}
\caption{t-SNE embeddings of Amazon (blue) and Webcam (green) images using our
supervised 256-dimensional representation learned with MMD regularization
(top left) and the original $fc7$ representation from the pre-trained model
(bottom right). Observe that the clusters formed by our representation separate
classes while mixing domains much more effectively than the original
representation that was not trained for domain invariance. For example, in
$fc7$-space the Amazon monitors and Webcam monitors are separated into distinct
clusters, whereas with our learned representation all monitors irrespective of
domain are mixed into the same cluster.}
\label{fig:tsne}
\end{figure*}

\begin{figure*}
\centering
\begin{subfigure}[b]{.48\linewidth}
\includegraphics[width=1.0\linewidth]{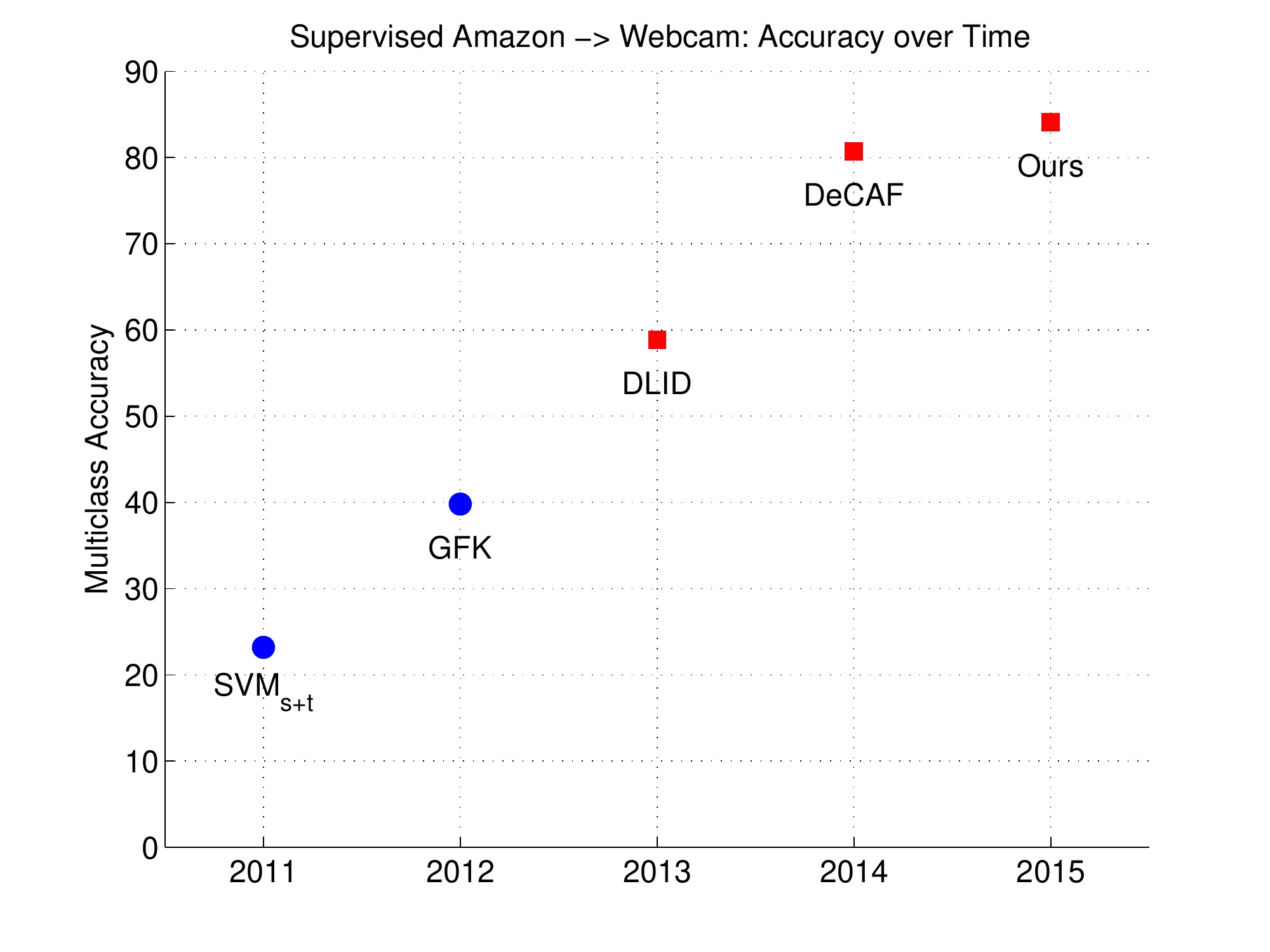}
\caption{A$\rightarrow$W supervised adaptation}
\end{subfigure}\hfill
\begin{subfigure}[b]{.48\linewidth}
\includegraphics[width=1.0\linewidth]{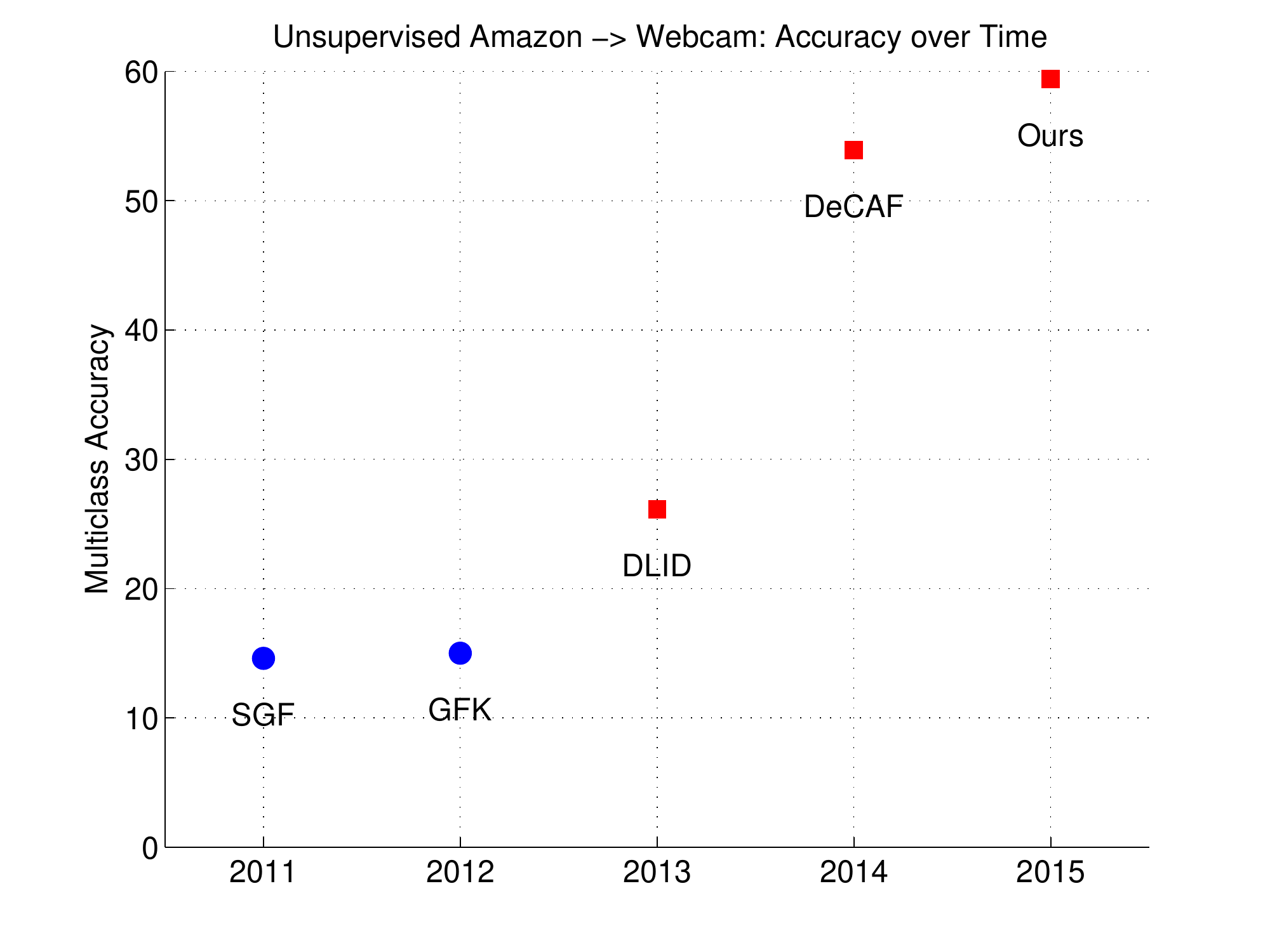}
\caption{A$\rightarrow$W unsupervised adaptation}
\end{subfigure}
\caption{Rapid progress over the last few years on a standard visual domain
adaptation dataset, \emph{Office}~\cite{saenko-eccv10}. We show methods on
Amazon$\rightarrow$Webcam that use traditional hand designed visual
representations with blue circles and methods that use deep representations are
depicted with red squares. For the supervised task, our method achieves 84\%
multiclass accuracy, an increase of 3\%.  For the unsupervised task, our method
achieves 60\% multiclass accuracy, an increase of 6\%.}
\label{fig:historical}
\end{figure*}

In order to determine how MMD regularization affects learning, we also compare
the learning curves with and without regularization on the
Amazon$\rightarrow$Webcam transfer task in Figure~\ref{fig:learning-curve}.  We
see that, although the unregularized version is initially faster to train, it
quickly begins overfitting, and test accuracy suffers. In contrast, using MMD
regularization prevents the network from overfitting to the source data, and
although training takes longer, the regularization results in a higher final
test accuracy.

To further demonstrate the domain invariance of our learned representation, we
plot in Figure~\ref{fig:tsne} a t-SNE embedding of Amazon and Webcam images
using our learned representation and compare it to an embedding created with
$fc7$ in the pretrained model. Examining the embeddings, we see that our learned
representation exhibits tighter class clustering while mixing the domains within
each cluster. While there is weak clustering in the $fc7$ embedding, we find
that most tight clusters consist of data points from one domain or the other,
but rarely both.

\subsection{Historical Progress on the Office Dataset}
In Figure~\ref{fig:historical} we report historical progress on the standard
Office dataset since it's introduction. We indicate methods which use
traditional features (ex: SURF BoW) with a blue circle and methods which use
deep representations with a red square. We show two adaptation scenarios.
The first scenario is a supervised adaptation task for visually distant domains
(Amazon$\rightarrow$Webcam). For this task our algorithm outperforms DeCAF by
3.4\% multiclass accuracy. Finally, we show the hardest task of unsupervised
adaptation for that same shift. Here we show that our method provides the most
significant improvement of 5.5\% multiclass accuracy.

\section{Conclusion}

In this paper, we presented an objective function for learning domain invariant
representations for classification. This objective makes use of an additional
domain confusion term to ensure that domains are indistinguishable in the
learned representation. We then presented a variety of ways to optimize this
objective, ranging from simple representation selection from a fixed pool to a
full convolutional architecture that directly minimizes the objective via
backpropagation.

Our full method, which uses MMD both to select the depth and width of the
architecture while using it as a regularizer during fine-tuning, achieves
state-of-the-art performance on the standard visual domain adaptation benchmark,
beating previous methods by a considerable margin.

These experiments show that incorporating a domain confusion term into the
discriminative representation learning process is an effective way to ensure
that the learned representation is both useful for classification and invariant
to domain shifts.

\section*{Acknowledgments}
This work was supported in part by DARPA's MSEE and SMISC programs, NSF awards IIS-1427425, IIS-1212798, and IIS-1116411, Toyota, and the Berkeley Vision and Learning Center.

{\small
\bibliographystyle{ieee}
\bibliography{main}
}

\end{document}